\documentclass[runningheads]{llncs}

\usepackage{eccv}

\usepackage{eccvabbrv}

\usepackage{graphicx}
\usepackage{booktabs}
\usepackage{framed}
\usepackage{mdframed}
\usepackage{graphicx}
\usepackage{amsmath}
\usepackage{amssymb}
\usepackage{booktabs}
\usepackage{colortbl}
\usepackage{enumerate}
\usepackage{multirow}
\usepackage{color}
\usepackage{pifont}
\usepackage{bbding}

\newcommand{\bftab}[1]{{\fontseries{b}\selectfont#1}}
\newcommand{\tabincell}[2]{\begin{tabular}{@{}#1@{}}#2\end{tabular}}

\usepackage[accsupp]{axessibility}  

\usepackage{hyperref}

\usepackage{orcidlink}

\begin{document}

\title{Occluded Gait Recognition with Mixture of Experts: An Action Detection Perspective}

\titlerunning{GaitMoE}

\author{Panjian Huang\inst{1,3}$^\star$\orcidlink{0000-0002-9777-4365} \and
Yunjie Peng\inst{2,4}\thanks{Equal contribution}\orcidlink{0000-0002-9275-0356} \and
Saihui Hou\inst{1,3}$^\dagger$\orcidlink{0000-0003-4689-2860} \and
Chunshui Cao\inst{3}\orcidlink{0000-0001-6634-1682} \and
Xu Liu\inst{3}\orcidlink{0000-0002-0401-1343} \and
Zhiqiang He\inst{2,4}\orcidlink{0000-0003-3103-1902} \and
Yongzhen Huang\inst{1,3}$^\dagger$\orcidlink{0000-0003-4389-9805}
}

\authorrunning{P. Huang et al.}

\institute{School of Artificial Intelligence, Beijing Normal University \and
School of Computer Science and Technology, Beihang University \and
WATRIX.AI \and
AI Lab, Lenovo Research
}

\maketitle

\renewcommand{\thefootnote}{\fnsymbol{footnote}}
\footnotetext{Corresponding author}

\begin{abstract}
Extensive occlusions in real-world scenarios pose challenges to gait recognition due to missing and noisy information, as well as body misalignment in position and scale.
We argue that rich dynamic contextual information within a gait sequence inherently possesses occlusion-solving traits:
1) Adjacent frames with gait continuity allow holistic body regions to infer occluded body regions; 
2) Gait cycles allow information integration between holistic actions and occluded actions.
Therefore, we introduce an action detection perspective where a gait sequence is regarded as a composition of actions. 
To detect accurate actions under complex occlusion scenarios, we propose an Action Detection Based Mixture of Experts (GaitMoE), consisting of Mixture of Temporal Experts (MTE) and Mixture of Action Experts (MAE).
MTE adaptively constructs action anchors by temporal experts and MAE adaptively constructs action proposals from action anchors by action experts.
Especially, action detection as a proxy task with gait recognition is an end-to-end joint training only with ID labels.
In addition, due to the lack of a unified occluded benchmark, we construct a pioneering Occluded Gait database (OccGait), containing rich occlusion scenarios and annotations of occlusion types.  
Extensive experiments on OccGait, OccCASIA-B, Gait3D and GREW demonstrate the superior performance of GaitMoE.
OccGait is available at \href{https://github.com/BNU-IVC/OccGait} {https://github.com/BNU-IVC/OccGait}.
\end{abstract}

\section{Introduction}
\begin{figure}[tbp]
  \centering
	\subfloat[Occlusion Issues\label{occlusion_issues}]{\includegraphics[width=.98\linewidth]{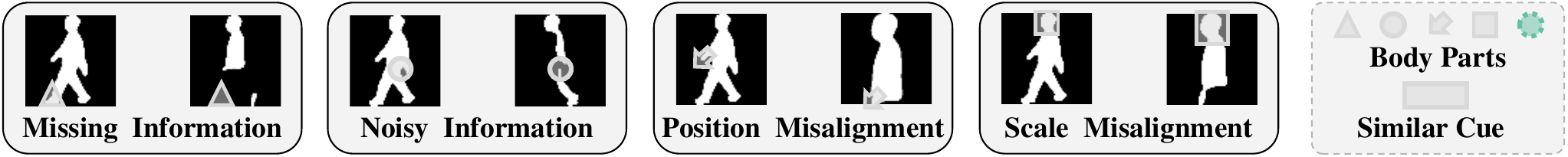}} \\
	\subfloat[Dynamic Contextual Information (Ours)\label{dynamic_context_information}]{\includegraphics[width=.98\linewidth]{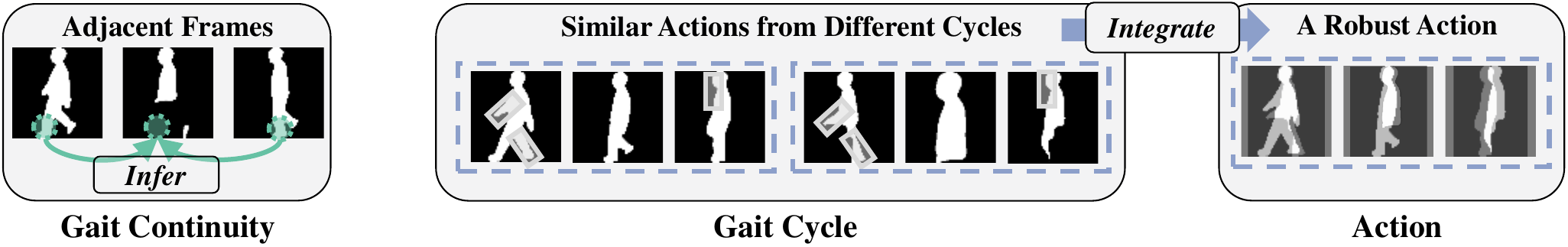}}
  \caption{(a) The main four occlusion issues in gait recognition. (b) The dynamic contextual information within a gait sequence can infer and integrate occlusion information.}
  \label{occlusion_solution}
\end{figure}

Gait recognition has attracted increasing attention and gained broad applications in crime
prevention, forensic identification, and social security~\cite{song2022casia} due to its ability to accurately identify walking patterns of pedestrians from a distance in complex surveillance scenarios, \eg, viewing angles, cloth-changing and illumination conditions~\cite{sepas2022deep}.
Applying upstream tasks such as tracking, segmentation, size normalization, and alignment to preprocess raw videos, the obtained gait representations (\eg, silhouettes or skeletons) make existing methods~\cite{chao2019gaitset,fan2020gaitpart, hou2020gait, lin2020gait, hou2021set, lin2021gait, chai2022lagrange, ma2023dynamic,dou2023gaitgci, huang2021context, wang2023hierarchical} achieve accurate identification.
However, current studies overlook occlusions largely existing in practical scenarios, \eg, occluded by carrying, obstacles, the crowd, or moving out of camera view. As shown in Fig.~\ref{occlusion_solution}(a), body regions occluded by obstacles or the crowd lead to missing and noisy information, while partial visual body regions with size normalization cause position and scale misalignment, which significantly degrading fine-grained feature matching~\cite{xu2023occluded, xu2023occlusion}.
Direct solutions, \eg, simply discarding or persisting occluded frames, do not adequately address occlusion issues since partially visual body regions in occluded frames may still contain key discriminative regions while allowing them to persist will cause erroneous feature extraction and matching.
Therefore, occlusion issues have become one of the biggest bottlenecks in gait recognition. 
%

To address occlusion issues in gait recognition, we rethink the dynamic contextual information within a gait sequence:
\textbf{(i) Gait Continuity.} Adjacent frames with continual motion enable body regions in holistic frames to infer the same body regions in occluded frames. As shown in Fig.~\ref{occlusion_solution}(b), for the current frame missing lower body information, the preceding and subsequent holistic frames can still infer approximate motion in the occluded regions.
\textbf{(ii) Gait Cycle.} As shown in Fig.~\ref{occlusion_solution}(b), we regard the current misaligned frame and adjacent frames as an action. Combined with the gait cycle, \ie, a gait sequence is formed by the repetition of a series of actions, to discover actions with similar cues, holistic and occluded actions can be integrated into a robust action.

Driven by the above analysis, we introduce a new perspective for occluded gait recognition, ``Action Detection''. 
The paradigm of action detection aims to predict the action boundaries and categories from an untrimmed video, where predefined consecutive frames as action anchors represent potential actions and further action anchors with high actionness scores generate action proposals~\cite{shi2023tridet, lin2021learning, yang2020revisiting}.
Specific to gait recognition, we associate action anchors to represent adjacent frames with gait continuity and further consider the gait cycle to construct action proposals from similar action anchors.
Therefore, a gait sequence can be regarded as a composition of actions.
%
\begin{figure}[tbp]
	\centering
	\includegraphics[width=1.0\linewidth]{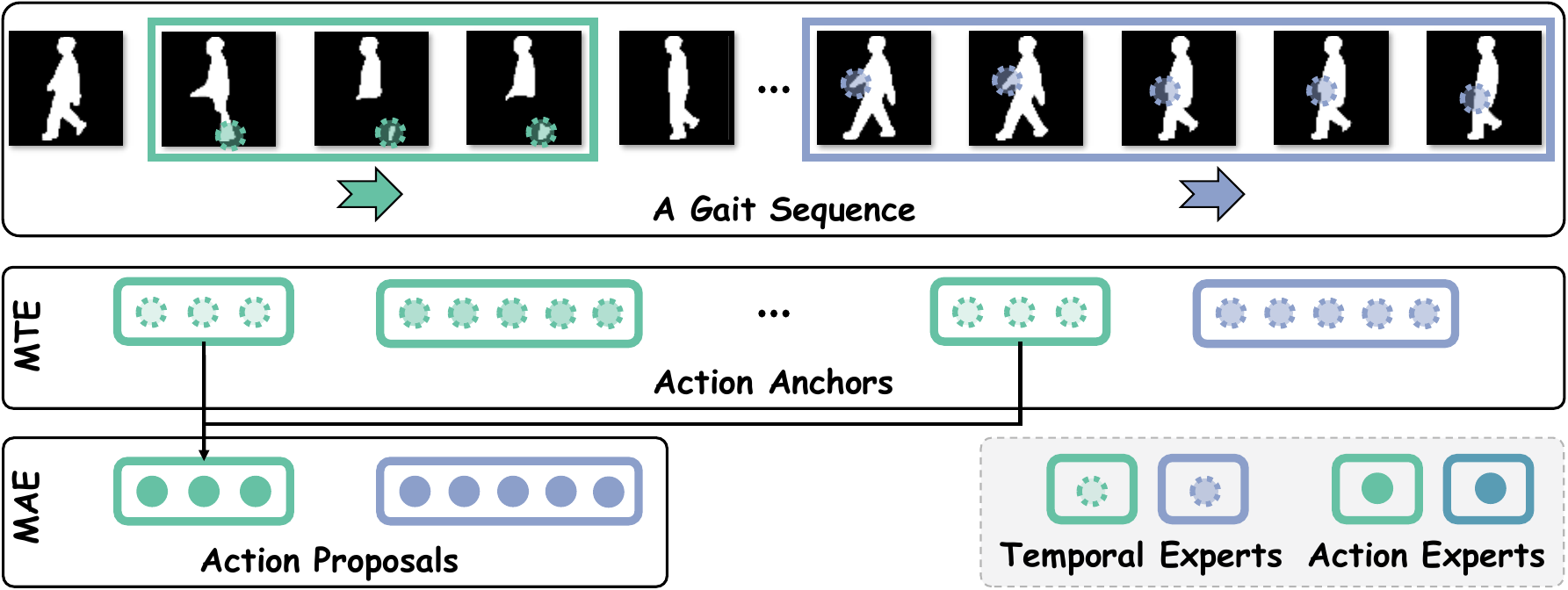}
	\caption{
    Action composition. Each temporal expert focuses on one body region with individual temporal size, constructing action anchors. Each action expert integrates similar action anchors from different gait cycles, constructing one action proposal.
	}
	\label{Gait_Action}
\end{figure}
%

Considering a single model that struggles to capture holistic and diverse actions under complex occlusion scenarios, \eg, the uncertainty of occluded body regions and duration, we propose an Action Detection Based Mixture of Experts (GaitMoE).
Mixture of Experts (MoE) follows a divide-and-conquer philosophy, breaking down complex problems into simple sub-problems.
Each sub-problem is handled by a dedicated expert, contributing collectively to solve the overall complexity.
As shown in Fig.~\ref{Gait_Action}, each temporal expert focuses on the corresponding body region and temporal granularity, sliding at the entire gait sequence to construct action anchors. 
Subsequently, each action expert integrates similar action anchors with one action type, constructing action proposals.
Finally, instead of localization and classification in action detection, action proposals are collectively as discriminative features for identification.

Additionally, the absence of publicly available gait databases with quantifiable occlusion metrics poses an extreme challenge for occluded gait recognition.
To this end, we establish a pioneering Occluded Gait database (OccGait) with two characteristics:
\textbf{(i) Diverse Occlusion Scenarios.} Each subject has $4$ different types of occlusion situations, including None of Occlusion, Carrying Occlusion, Crowd Occlusion, and Static Occlusion.
\textbf{(ii) Explicit Occlusion Types.} OccGait provides explicit occlusion types for each gait sequence, which enables to qualify and quantify occlusion issues.

Our main contributions can be summarized as follows: 
\begin{itemize}
    \item To address occlusion challenges, we introduce an action detection perspective where an Action Detection Based Mixture of Experts (GaitMoE) structures a gait sequence as a composition of action.
    \item To qualify and quantify occlusion issues, we build a novel Occluded Gait recognition benchmark (OccGait), including diverse occlusion scenarios and explicit annotations of occlusion types.
    \item To evaluate effectiveness and robustness, extensive experimental results on OccGait, OccCASIA-B, Gait3D, and GREW demonstrate that our method significantly outperforms other state-of-the-art methods.
\end{itemize}

\section{Related Work}
\subsection{Gait Recognition}
Gait Recognition is mainly categorized into appearance-based and model-based approaches.
Appearance-based methods~\cite{chao2019gaitset,fan2020gaitpart, hou2020gait, lin2020gait, hou2021set, lin2021gait, chai2022lagrange, ma2023dynamic,dou2023gaitgci, huang2021context, wang2023hierarchical} usually uses templates of compressing a sequence of gait silhouettes (\eg, Gait Energy Image), set of frames and sequence of frames as inputs, extracting fine-grained features (\eg, spatial-temporal and part-level representations).
Model-based methods~\cite{teepe2021gaitgraph, teepe2022towards, zhang2023spatial, fu2023gpgait, zheng2022gait, zhu2023gaitref} explicitly model human body structure, \eg, 2D or 3D skeletons and meshes. 
Additionally, some researches~\cite{liang2022gaitedge, shen2023lidargait, bashir2009gait} take other data types as inputs, such as RGB frames, optical flow and point clouds. 
However, most of these methods usually neglect the fact that real-world scenarios introduce a significant amount of occlusion.

\subsection{Occluded Gait Recognition} 
We introduce occluded gait recognition from two aspects: \textbf{(i) DataBases.} For synthesis-based databases, Chen \emph{et al.}~\cite{chen2009frame} simulate occluded scenarios based on CMU Mobo~\cite{gross2001cmu} through adding horizontal or vertical black bars. 
Uddin \emph{et al.}~\cite{uddin2019spatio} synthesize relative static and dynamic occlusions based on OUMVLP~\cite{takemura2018multi} by a background rectangle mask in a fixed position or gradually changed position.
Delgado-Escano \emph{et al.}~\cite{delgado2020mupeg} generate crowd occlusions based on CASIA-B~\cite{yu2006framework} and TUM-GAID~\cite{hofmann2014tum} by augmenting persons in raw videos.
Xu \emph{et al.}~\cite{xu2023occlusion, xu2023occluded} synthesize occluded scenarios by simulating cropping and size-normalized silhouettes.
For real-world databases, Hofmann \emph{et al.}~\cite{hofmann2011identification} collect TUM-IITKGP, including static occlusions (\eg, standing people, a backpack, gown and hands in pocket), dynamic occlusions (\eg, two walking people).
Chattopadhyay \emph{et al.}~\cite{chattopadhyay2015frontal} construct a frontal and occluded gait database by estimating Kinect depth.
Li \emph{et al.}~\cite{li2023multi} present an OG RGB+D dataset captured by Azure Kinect DK sensor, containing occlusions with carrying, clothing and the crowd.
However, these databases have some limitations, \eg, the single occlusion scenario, small occlusion regions, or not publicly available yet.
\textbf{(ii) Architectures.} For reconstruction-based approaches, Xu \emph{et al.}~\cite{xu2023occluded} re-normalize and register silhouettes from learned holistic information (\eg, scales) before the following feature extraction and matching process.
Xu \emph{et al.}~\cite{xu2023occlusion} estimate SMPL with pose and shape features from RGB occluded videos.
Uddin \emph{et al.}~\cite{uddin2019spatio} reconstruct a sequence from the occluded sequence by a conditional deep generative adversarial network.
Peng \emph{et al.}~\cite{peng2023occluded} register and recover occluded silhouettes with a self-supervised alignment module and temporal recovery transformer. 
Guo \emph{et al.}~\cite{guo2023physics} propose a Physics-Augmented Autoencoder (PAA) that generates physically intermediate representations through a graph-convolution-based encoder and a physics-based decoder, which enhances the ability to reconstruct input skeleton sequences even when partially occluded.
For reconstruction-free approaches, Gupta \emph{et al.}~\cite{gupta2024you} propose a occlusion-aware module by synthetic occlusions to detect occlusion type information to guide gait recognition training. 
Zhu \emph{et al.}~\cite{zhu2023gait} use SMPLify-X that provides the body shape feature decoupled from its pose and strong prior, which enables to generate the complete shape even with mild occlusions.

\subsection{Mixture of Experts}
Mixture of Experts (MoE) is a sparsely-activated architecture where a router network output weights for aggregating multiple experts~\cite{shazeer2017outrageously}. 
The philosophy of divide and conquer allows MoE to reduce computational cost and increase model capacity, and it has been widely extended to Vision Transformer~\cite{riquelme2021scaling, wang2022image, li2022sparse}. 
Each expert in MoE is dispatched with one sub-data (\eg, image patches, data from one domain) and maintains specialization, which is named ``Expert''. 
This work extends MoE to detect fine-grained actions in occluded gait recognition and makes each expert concentrate on one representative action.

\section{Methodology}
\begin{figure}[tbp]
	\centering
	\includegraphics[width=1.0\linewidth]{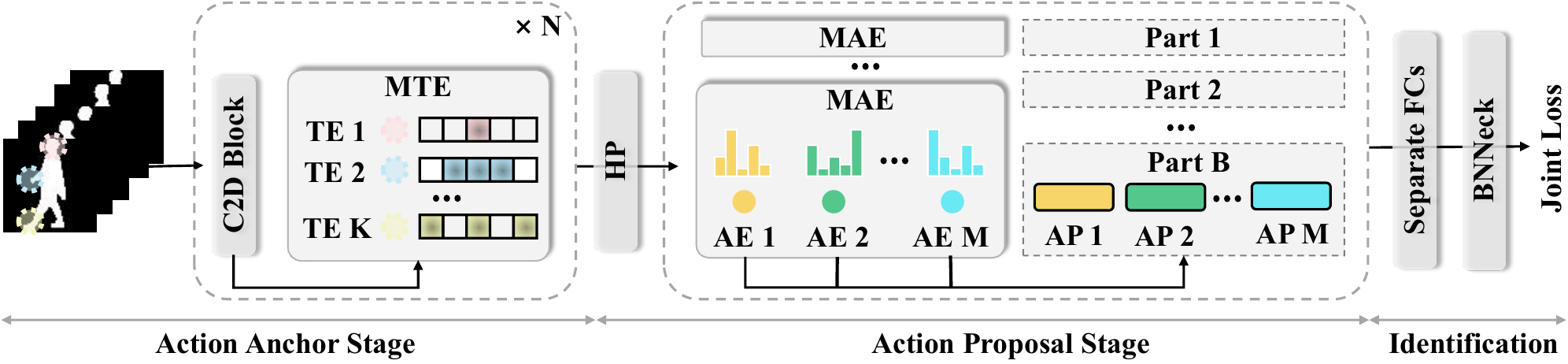}
	\caption{
    The overview of GaitMoE. Best viewed in colors, the input sequence is firstly extracted to frame-level features by C2D Block (\eg, 2D CNNs or a residual block), and Mixture of Temporal Experts (MTE) dispatch temporal experts (TE) with different temporal bounding boxes to the corresponding channel segments, forming action anchors. After horizontal partitioning (HP), Mixture of Action Experts (MAE) is independent for each part-level feature. For each channel segment, one action expert (AE) integrates similar action anchors along the temporal dimension by weighted sum operation, forming one action proposal. Finally, The concatenated action proposals as part features for identification.
	}
	\label{Overview_GaitMoe}
\end{figure}

GaitMoE mainly consists of Mixture of Temporal Experts (MTE) and Mixture of Action Experts (MAE).
We give a brief overview of the full process in Fig.~\ref{Overview_GaitMoe}.
%

\subsection{Mixture of Temporal Experts}
Although a gait sequence has filtered texture information and performed coarse-aligned registration, the complex occlusions in real scenarios cause the uncertainty of occluded body regions and duration.
To this end, we adopt multi-scale mechanisms in temporal and channel dimensions to extract fine-grained features.

%
\begin{figure}[tbp]
  \centering
	\subfloat[MTE\label{MTE}]{\includegraphics[width=.48\linewidth]{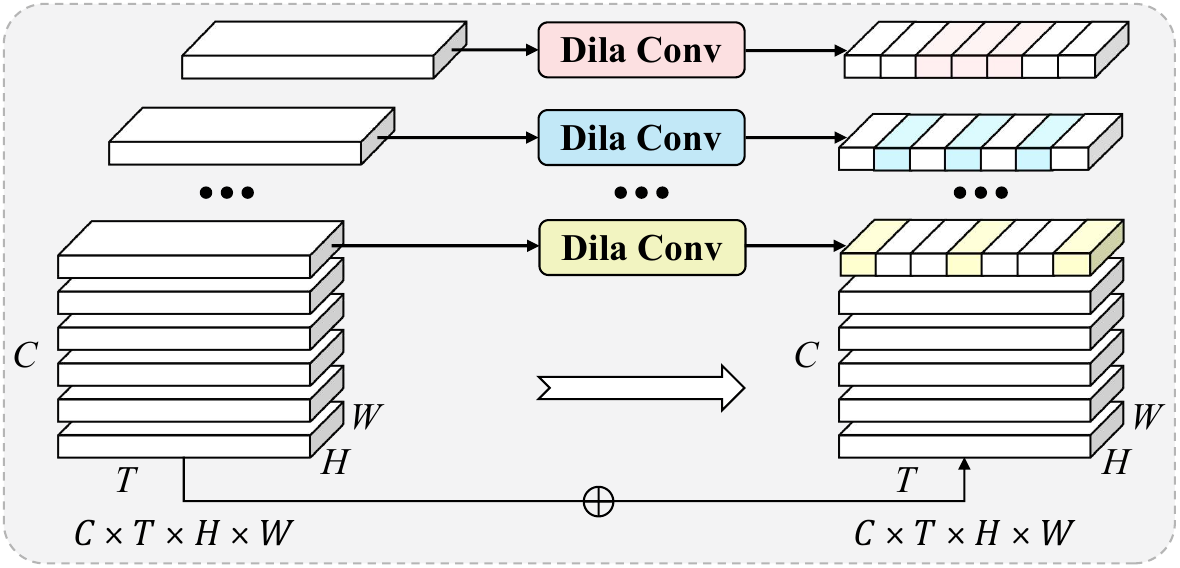}}\hspace{5pt}
	\subfloat[MAE\label{MAE}]{\includegraphics[width=.48\linewidth]{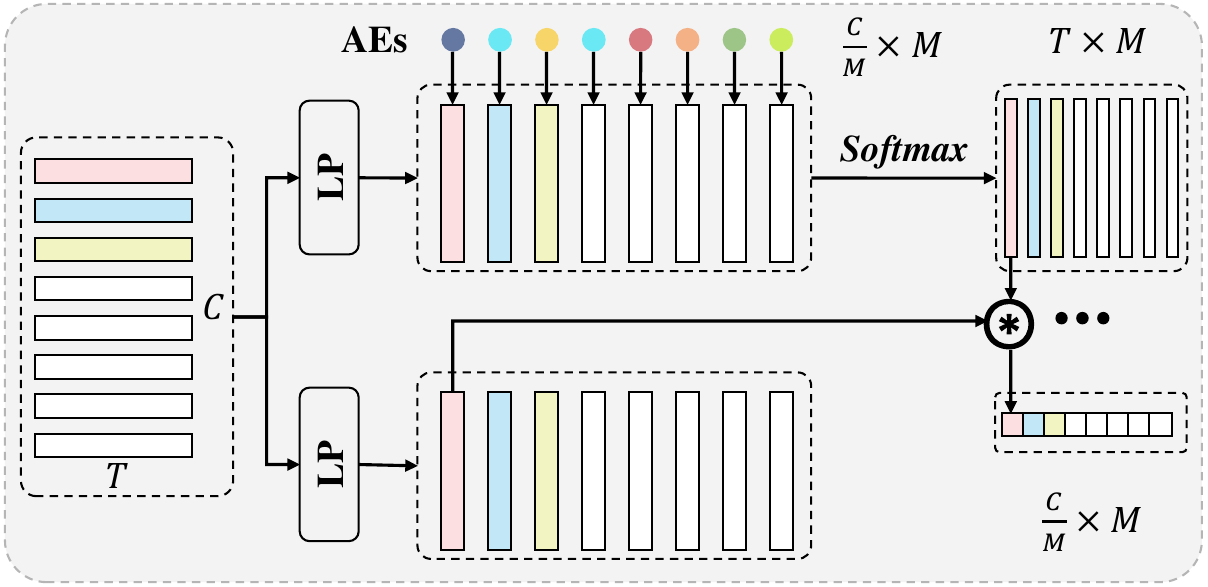}}
  \caption{(a) Mixture of Temporal Experts. Dila Conv (DC) represents Dilated Convolution, predefining action anchors with different dilated ratios. (b) Mixture of Action Experts. LP denotes the linear projection. Similar action anchors adaptively integrate into action proposals.}
  \label{mte_mae}
\end{figure}

\noindent \textbf{Action Anchors.}
As we know, a gait sequence possesses continuity with significant mutual information between each frame and adjacent frames. 
For example, when we observe a person lifting their leg, it is likely to be followed by a swinging leg. 
Some existing approaches employ temporal modeling to capture such relationships, \eg, 3D CNNs and LSTMs. 
However, uncertain and complex occlusions interfere with the dynamic information. 
To alleviate these issues, Fig.~\ref{mte_mae}(a) shows that MTE predefines various sizes of temporal experts, which are dilated convolutions with different dilated ratios for corresponding channel segments. 
At each temporal position, each temporal expert independently constructs action anchors from adjacent temporal positions, which is why we name it ``Temporal Expert''.
Considering adjacent occluded frames may introduce noise information to current holistic frames, MTE preserves partial channel segments for adaptively selecting clean regions.
Let $\mathcal X \in \mathbb{R}^{\mathcal C \times \mathcal T \times \mathcal H \times \mathcal W}$ denote frame-level features extracted from silhouettes by C2D Block, where $\mathcal C, \mathcal T, \mathcal H, \mathcal W$ represent channel, consecutive $\mathcal T$ frames, height and width dimensions. 
The process of MTE is formulated as follows:
\begin{equation}
\begin{aligned}
\mathcal Y = Concat (DC _{i}(\mathcal X _{i}), i =1, 2,\cdots, \mathcal K, \mathcal X _{[\mathcal S - \mathcal K + 1, \mathcal S]})
\end{aligned}
\end{equation}
where $\mathcal X _{i} \in \mathbb{R}^{\frac{\mathcal C}{\mathcal S} \times \mathcal T \times \mathcal H \times \mathcal W}$, $\mathcal Y \in \mathbb{R}^{\mathcal C \times \mathcal T \times \mathcal H \times \mathcal W}$, $\mathcal S$ is the number of channel segments, $\mathcal K$ is the number of temporal experts, $DC$ is dilated convolution, and $i$ is the segment index. 
In our work, we set $\mathcal S =8$, $\mathcal K =4$, and $DC_{i}$ is 3D CNN with kernel size $(3, 1, 1)$, stride $(1, 1, 1)$, padding $(i, 0, 0)$, and dilated ratio $i$. In addition, residual learning is embedded within MTE for easing training.

\subsection{Mixture of Action Experts}
Although action anchors contain rich dynamic information, they may have a large amount of redundant and invalid actions.
Therefore, we adopt prototype-based architecture to adaptively select and integrate discriminative and similar action anchors as action proposals.
Since a gait sequence can be regarded as the composition of actions, when the sequence is occluded resulting in many invalid actions, the action prototype needs to detect the most discriminative action type and aggregate this action type from occluded and holistic actions. 
In addition, GaitMoE adopts horizontal pooling (HP) for fine-grained part features $\mathcal P \in \mathbb{R}^{\mathcal C \times \mathcal T}$, and MAE is independent for each part.
Here, we omit the part index for simplicity.

\noindent \textbf{Action Proposals.} To capture discriminative actions only with ID labels, MAE shown in Fig.~\ref{mte_mae}(b) predefines a set of learnable action prototypes where an action prototype adaptively learns a type of action for recognition, that is why we name it ``Action Expert''.
For fine-grained action extraction, we dispatch each action expert to the corresponding channel segment (\eg, action anchors), and action experts ``watch'' contextual action anchors in the entire temporal dimension for filtering action anchors with occlusions, and select and integrate similar action anchors as action proposals.
Different to MoEs~\cite{shazeer2017outrageously, mustafa2022multimodal, lepikhin2020gshard, fedus2022switch} where the routers generally adopt Top-K selection and sparsely memorize information, recent MoE works has shown remarkable performance with a fixed hash router~\cite{roller2021hash}, or convolutional experts~\cite{chen2023learning}. 
In this work, we introduce a soft selection for balancing the training.
Let $\mathcal A \in \mathbb{R}^{\frac{\mathcal C}{\mathcal M} \times \mathcal M}$ represent $\mathcal M$ action experts with $\frac{\mathcal C}{\mathcal M}$ dimension, and we obtain action queries $\mathcal Q$ by identify mapping on $\mathcal A$, action keys $\mathcal K$ and values $\mathcal V$ by different linear projections on $\mathcal P$.
Then, we dispatch each action query to the corresponding channel segment of $\mathcal K$ to evaluate action anchors by scores where the higher the score, the more reliable the action anchor, and vice versa. 
To make one action expert concentrate on one most representative action, we use the softmax function to calculate the scores within the corresponding channel segment of $\mathcal K$ and weighted sum operation with the corresponding channel segment of $\mathcal V$ as an action proposal.
The formulation is as follows:
\begin{equation}
\begin{aligned}
\mathcal Q _{i} = \mathcal A _{i},\quad \mathcal K _{i} = \mathcal P_{i} \mathcal W ^{\mathcal K},\quad \mathcal V _{i} = \mathcal P_{i} \mathcal W ^{\mathcal V} 
\end{aligned}
\end{equation}
\begin{equation}
\begin{aligned}
\mathcal F _{i} = \sum\limits_{j=1}^{\mathcal T} \mathcal O_{i, j} \otimes \mathcal V _{i, j},\quad \mathcal O_{i, j} = \frac{exp(\mathcal Q _{i, j}\mathcal K _{i, j}^{T})}{\sum_{j=1}^{\mathcal T} exp(\mathcal Q _{i, j}\mathcal K _{i, j}^{T})}
\end{aligned}
\end{equation}
where $i$ is the channel segment index, $i \in 1, 2,\cdots, \mathcal M$,
$j$ is the action anchor index along the temporal dimension of the $i$ segment, $j \in 1, 2,\cdots, \mathcal T$, $\mathcal W ^{\mathcal K}, \mathcal W ^{\mathcal V} \in \mathbb{R}^{\frac{\mathcal C}{\mathcal M} \times \frac{\mathcal C}{\mathcal M}}$, $\mathcal P_{i} \in \mathbb{R}^{\mathcal T \times \frac{\mathcal C}{\mathcal M}}$.
$\mathcal Q _{i, j}, \mathcal K _{i, j}, \mathcal V _{i, j} \in \mathbb{R}^{1 \times\frac{\mathcal C}{\mathcal M}}$.
Finally, we obtain $\mathcal F$ as one part feature by concatenating action proposals $[\mathcal F _{1}, \mathcal F _{2},\cdots, \mathcal F _{\mathcal M}]$ along the channel dimension.
$\mathcal F$ is fed into Separate FCs and BNNeck for identification.

\subsection{Joint Loss}
GaitMoE is an end-to-end joint learning framework only with ID labels, introducing action detection as a proxy task with gait recognition. 
The joint loss includes two types: Triplet Loss~\cite{hermans2017defense} $\mathcal L _{tp}$ and Cross Entropy Loss $\mathcal L _{ce}$, which constrains each part independently. This formulation is as follows:
\begin{equation}
\begin{aligned}
\mathcal L = \mathcal L _{tp} + \beta \mathcal L _{ce}
\end{aligned}
\end{equation}
where the hyper-parameter $\beta$ is for balancing the two terms.
%


\section{The OccGait Benchmark}
\label{section4}
Due to the absence of a comprehensive gait database for quantitatively and qualitatively analyzing the impact of various occlusion types, we collect the Occluded Gait Database (OccGait), including $101$ subjects with $4$ types of occluded scenarios, $8$ camera views, and over $80k$ sequences. 
\textit{It is worth noting that we hope the OccGait can serve as a starting point to promote robust gait recognition for practical applications, similar to the transition from the indoor databases, \eg, CASIA-X Series~\cite{wang2003silhouette, yu2006framework, tan2006efficient, song2022casia} and OU-X Series~\cite{an2020performance, takemura2018multi, uddin2018isir, xu2017isir, iwama2012isir, hossain2010clothing, makihara2011gait, tsuji2010silhouette} to the wild databases, \eg, Gait3D~\cite{zheng2022gait} and GREW~\cite{zhu2021gait}}.

\subsection{Data Collection and Pre-processing}
\begin{figure}[tbp]
	\centering
	\subfloat[Layout and camera views]{
		\label{NM}\includegraphics[width=.45\linewidth]{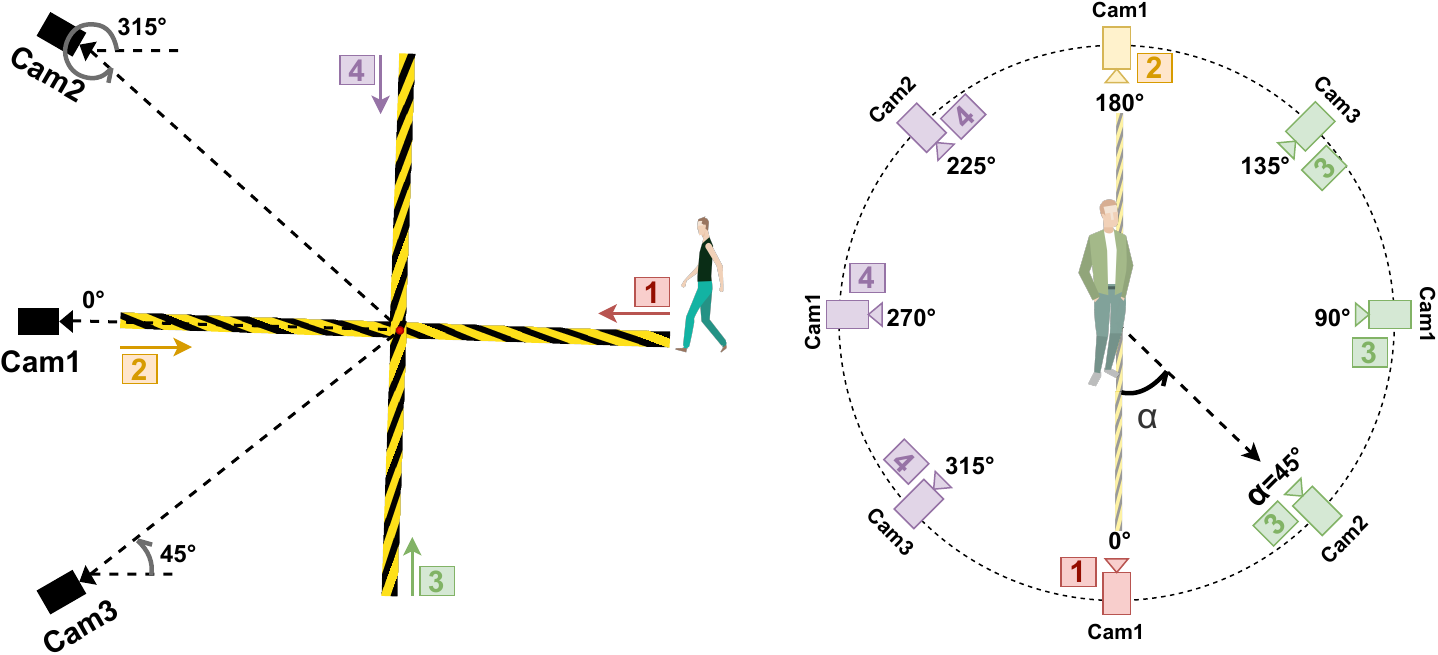}
	}
	\subfloat[Occlusions]{
		\label{Occlusion}\includegraphics[width=.45\linewidth]{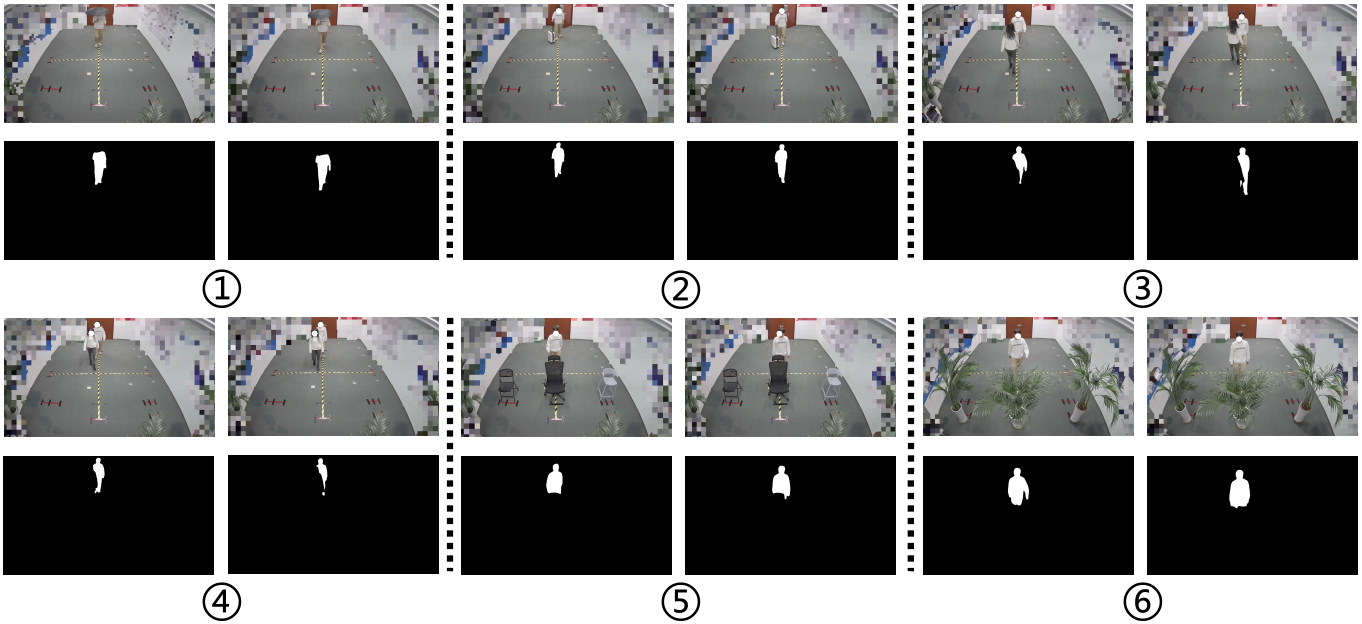}
	}
	\caption{
        (a) The layout and camera views during collection. (b) The $3$ types of occlusion scenarios where the number \textcircled{1}\textcircled{2} are for Carrying Occlusion, \textcircled{3}\textcircled{4} are for Crowd Occlusion and \textcircled{5}\textcircled{6} are for Static Occlusion, and None Occlusion is omitted for simplicity.
	}
	\label{OccGait}
\end{figure}

The OccGait is collected in an indoor gait recognition laboratory. 
During OccGait collection, we obtain authorization from all subjects who are informed for academic data collection in advance. Privacy is also the highest priority in our research.
As the left in Fig.~\ref{OccGait}(a), we place $3$ cameras (Cam1 of $0^{\circ}$, Cam2 of $45^{\circ}$, Cam3 of $315^{\circ}$) with $1920 \times 1080$ resolution in the square area.
During the data collection process, the subjects follow this walk route (\ie, 1-2-3-4). We filter out data with overlapping camera views caused by the combination of 3 cameras and walking directions, and obtain gait sequences with $8$ camera views on the right in Fig.~\ref{OccGait}(a). 
To qualify and quantify realistic and complex occluded scenarios, we set $4$ types of occluded situations with diverse occlusions shown in Fig.~\ref{OccGait}(b), which are None of Occlusion (\ie, Normal Walking as NM), Carrying Occlusion (CA, \textcircled{1}\textcircled{2}), Crowd Occlusion (CR, \textcircled{3}\textcircled{4}), and Static Occlusion (ST, \textcircled{5}\textcircled{6}).
%

All subjects walk the route four times for NM and two times for CA, CR and ST, respectively, which denotes NM$01$, NM$02$, NM$03$, NM$04$, CA$01$, CA$02$, CR$01$, CR$02$, ST$01$ and ST$02$.

\noindent \textbf{None of Occlusion.} 
To simulate walking status in real scenarios, subjects in their clothing, walk with their walking speed.
As shown in Fig.~\ref{OccGait}(a), there are no obstacles in this situation, and the full body of each subject is fully visible.

\noindent \textbf{Carrying Occlusion.} 
As shown in Fig.~\ref{OccGait}(b)(\textcircled{1}\textcircled{2}), we establish two common carrying scenarios for daily life: umbrellas and luggage. 
People with an umbrella on a rainy day partially obstruct the upper of the body, while luggage occludes both the torso and the lower of the body.

\noindent \textbf{Crowd Occlusion.} Gait recognition often requires retrieval in crowded scenes, and human body occlusion can lead to significant interference, such as occlusion of body edges and incorrect dynamic information, as shown in Fig.~\ref{OccGait}(b)(\textcircled{3}\textcircled{4}). 
We design two types of crowd occlusion, the subject walking with another person in different directions (opposite and parallel), which results in partial occlusion of gait sequences at certain moments and complete occlusion of gait sequences at all times.

\noindent \textbf{Static Occlusion.} Gait recognition will be deployed in a wide range of scenarios, such as in squares, malls, and other locations with numerous static obstructions. 
Fig.~\ref{OccGait}(b)(\textcircled{5}) demonstrates our placement of plants and chairs in the walking route.
Fig.~\ref{OccGait}(b)(\textcircled{6}) shows that the complex and irregular static obstructions hinder the lower body region.

\noindent \textbf{Data Pre-processing.} 
We adopt MaskFormer~\cite{cheng2022masked}, a segmentation algorithm pre-trained on large datasets (including numerous occluded scenarios), to extract silhouettes from the original RGB data as input.

\subsection{Evaluation Protocol}
OccGait is divided into training and testing sets. 
The $51$ individuals with odd numbers as the training set while the remaining $50$ with even numbers as the test set.
Our gait evaluation follows the protocols of the previous gait evaluation~\cite{yu2006framework}. 
Given a query sequence, we measure its distance to each sequence in the gallery, retrieving the subject with the closest distance from the gallery. 
To quantify occluded gait analysis, we use the NM01 and NM02 of each subject in the testing set as the gallery, evaluating their Rank 1 performance under different occlusions and viewing angles.

\section{Experiments}
\subsection{Datasets}
We first conduct extensive qualitative and quantitative occlusion analyses on OccGait and OccCASIA-B~\cite{peng2023occluded}.
Subsequently, we further validate the generalizability and practicality of our method on Gait3D~\cite{zheng2022gait} and GREW~\cite{zhu2021gait}. 

\noindent \textbf{OccGait} is for real-scenario occlusion evaluation built by this work. The details have been discussed in Section~\ref{section4}.

\noindent \textbf{OccCASIA-B} is a synthetic occluded gait database~\cite{peng2023occluded} from CASIA-B and has similar basic statistics, containing $124$ subjects, $3$ different walking situations, \etc, Walking in Normal (NM), Walking with a Bag (BG) and Walking with Different Clothes (CL), $11$ camera views from uniform interval of $18^{\circ}$ in $[0^{\circ}, 180^{\circ}]$. 
To simulate occlusion situations, OccCASIA-B sets $4$ types of occlusions: None Occlusion (NO), Crowd Occlusion (CO), Static Occlusion (SO) and Detect Occlusion (DO), which denotes a walking person without occlusions, occluded by another one in a crowded area, occluded by static occlusions, \eg,  benches, bicycles and fire hydrants, and losing body regions in the up, down, left, or right direction.
The OccCASIA-B takes the first $74$ subjects as the training set where each gait sequence with $0.6$ of occlusion probability generates one of the $4$ types of occluded scenarios.
The remaining $50$ subjects are used for occlusion evaluation. For each occlusion benchmark, except for the first $4$ NM sequences used as the holistic gallery set, the remaining sequences are generated with the corresponding occlusion scenario. 

\noindent \textbf{Gait3D} samples two segments of continuous two-hour video clips from each of seven-day raw videos in a supermarket, including complex covariates (\eg, occlusions, view angles) for practical gait recognition. It contains 3000 subjects with 25309 sequences, taking 2000 subjects as the training dataset and 1000 subjects as the testing dataset.

\noindent \textbf{GREW} is a large-scale wild gait database, containing 26345 subjects with 128671 sequences captured by 882 cameras. It provides 4 types of silhouettes, optical flow, and 2D/3D human poses. The benchmark takes 20000 subjects as the training dataset and 6000 subjects as the testing dataset, and each subject provides two sequences for the gallery set and two sequences for the probe set.

\begin{table}[t]
\centering
\footnotesize
	\caption{
		The Rank-1 accuracy (\%) on OccGait for different probe views excluding the identical-view cases.
		For evaluation, the sequences of NM01 and NM02 for each subject are taken as the gallery. The benchmark adopts None Occlusion (NO), Carrying Occlusion (CA), Crowd Occlusion (CR) and Static Occlusion (ST).
	}
	\label{OccGait_Performance}
    \setlength{\tabcolsep}{3mm}
    \resizebox{\linewidth}{!}{
			\begin{tabular}{c|c|c|c|c|c|c|c|c|c|c}
                \hline
				\multirow{2}{*}{\tabincell{c}{}} & \multirow{2}{*}{Method} & \multicolumn{8}{c|}{Probe View} & \multirow{2}{*}{Average} \\
				\cline{3-10}
				& & $0^{\circ}$ & $45^{\circ}$ & $90^{\circ}$ & $135^{\circ}$ & $180^{\circ}$ & $225^{\circ}$
				& $270^{\circ}$ & $315^{\circ}$ & \\
                \hline
				\multirow{6}{*}{NM}
				& GaitSet~\cite{chao2019gaitset}  &65.7 &91.7 &89.1 &90.7 &66.7 &91.9 &89.4 &92.1 & 84.7\\
                & GaitPart~\cite{fan2020gaitpart}  &62.9 &92.0 &88.9 &89.6 &59.6 &89.9 &87.3 &90.7 &82.6\\
                & GaitGL~\cite{lin2021gait}  &73.9 &94.3 &92.6 &93.4 &68.1 &93.1 &91.7 &92.6 &87.5\\
                & STOR~\cite{peng2023occluded}  &73.7 &94.6 &92.0 &93.3 &73.3 &94.1 &92.6 &92.6 &88.3\\
                & GaitBase~\cite{fan2023opengait}  &68.4 &91.6 &88.7 &91.1 &74.9 &93.6 &88.1 &91.7 & 86.0\\
				& \cellcolor{gray!20}\bftab{GaitMoE(\bftab{ours})}                       & \cellcolor{gray!20}\bftab{81.0} & \cellcolor{gray!20}\bftab{96.0} & \cellcolor{gray!20}\bftab{94.0} & \cellcolor{gray!20}\bftab{95.1} & \cellcolor{gray!20}\bftab{81.3} & \cellcolor{gray!20}\bftab{94.7} & \cellcolor{gray!20}\bftab{94.1} & \cellcolor{gray!20}\bftab{94.7} & \cellcolor{gray!20}\bftab{91.4}\\
				\hline
				\multirow{6}{*}{CA}
				& GaitSet~\cite{chao2019gaitset}  &50.1 &74.3 &79.7 &79.1 &47.4 &73.6 &74.0 &76.0 &69.3\\
                & GaitPart~\cite{fan2020gaitpart} &42.0 &69.9 &74.9 &78.6 &37.6 &66.4 &66.0 &63.7 & 62.4\\
                & GaitGL~\cite{lin2021gait}  &48.6 &79.7 &84.6 &87.9 &38.7 &76.7 &78.1 &70.4 &70.6\\
                & STOR~\cite{peng2023occluded}  &55.9 &83.6 &83.6 &86.3 &54.4 &81.7 &83.1 &81.7 &76.3\\
                & GaitBase~\cite{fan2023opengait}  &58.1 &82.0 &84.3 &85.6 &54.3 &79.9 &80.1 &79.3 & 75.4\\
				& \cellcolor{gray!20}\bftab{GaitMoE(\bftab{ours})}                       & \cellcolor{gray!20}\bftab{68.3} & \cellcolor{gray!20}\bftab{87.7} & \cellcolor{gray!20}\bftab{88.9} & \cellcolor{gray!20}\bftab{89.6} & \cellcolor{gray!20}\bftab{61.7} & \cellcolor{gray!20}\bftab{86.4} & \cellcolor{gray!20}\bftab{86.6} & \cellcolor{gray!20}\bftab{87.3} & \cellcolor{gray!20}\bftab{82.1}\\
				\hline
				\multirow{6}{*}{CR}
				& GaitSet~\cite{chao2019gaitset}  &58.3 &84.7 &80.1 &77.4 &52.3 &77.9 &77.6 &85.6 &74.2\\
                & GaitPart~\cite{fan2020gaitpart}  &48.1 &81.9 &76.4 &69.6 &39.0 &67.6 &68.1 &79.3 &66.3\\
                & GaitGL~\cite{lin2021gait}  &47.4 &89.0 &81.3 &77.1 &41.0 &75.0 &77.1 &87.6 &71.9\\
                & STOR~\cite{peng2023occluded}  &55.6 &88.4 &86.0 &81.7 &52.0 &81.7 &83.0 &88.4 &77.1\\
                & GaitBase~\cite{fan2023opengait}  &62.4 &86.9 &83.1 &82.1 &\bftab{60.0} &\bftab{85.4} &80.7 &87.6 &78.5\\
				& \cellcolor{gray!20}\bftab{GaitMoE(\bftab{ours})}                       & \cellcolor{gray!20}\bftab{63.1} & \cellcolor{gray!20}\bftab{90.6} & \cellcolor{gray!20}\bftab{86.6} & \cellcolor{gray!20}\bftab{84.4} & \cellcolor{gray!20}57.6 & \cellcolor{gray!20}81.6 & \cellcolor{gray!20}\bftab{84.9} & \cellcolor{gray!20}\bftab{90.3} & \cellcolor{gray!20}\bftab{79.9}\\
				\hline
				\multirow{6}{*}{ST}
				& GaitSet~\cite{chao2019gaitset}  &54.1 &86.3 &86.7 &82.3 &54.1 &86.7 &86.4 &83.6 &74.2\\
                & GaitPart~\cite{fan2020gaitpart}  &44.6 &83.7 &85.7 &77.3 &39.1 &83.4 &84.1 &77.4 &71.9\\
                & GaitGL~\cite{lin2021gait}  &36.7 &87.7 &90.7 &80.0 &37.6 &86.6 &90.4 &82.3 &74.0\\
                & STOR~\cite{peng2023occluded}  &50.3 &90.0 &91.3 &87.4 &54.9 &90.1 &91.3 &89.1 &80.6\\
                & GaitBase~\cite{fan2023opengait}  &57.7 &89.9 &87.4 &85.9 &59.9 &88.9 &87.4 &85.4 &80.3\\
				& \cellcolor{gray!20}\bftab{GaitMoE(\bftab{ours})}                      & \cellcolor{gray!20}\bftab{62.9} & \cellcolor{gray!20}\bftab{93.3} & \cellcolor{gray!20}\bftab{92.1} & \cellcolor{gray!20}\bftab{89.4} & \cellcolor{gray!20}\bftab{63.6} & \cellcolor{gray!20}\bftab{91.1} & \cellcolor{gray!20}\bftab{93.6} & \cellcolor{gray!20}\bftab{91.4} & \cellcolor{gray!20}\bftab{84.7}\\
				\hline
			\end{tabular}
		}
\end{table}

\subsection{Implementation Details}
We provide details about the training process. \textbf{Inputs.} All datasets are resized to $64 \times 44$. In addition, we employ spatial alignment module~\cite{peng2023occluded} as pre-processing to re-align input silhouettes for OccGait and OccCASIA-B. We adopt batch size [P, K] and the number of iterations, $[8, 16], 40K$ for OccGait, OccCASIA-B, $[32, 4], 60K$ for Gait3D, and $[32, 4], 180K$ for GREW.
We sample $30$ frames of each gait sequence in the training stage and all frames are used for inference. \textbf{Network.} For OccGait and OccCASIA-B, we stack three 2D convolution blocks as our Baseline with the number of channels (64, 128, 256). Each 2D convolution block is followed by an MTE. After Horizontal Pooling with the part parameter of 64, we set individual MAE for each part as in Fig.~\ref{mte_mae}(b) and $\mathcal M$ is set to 16. For Gait3D and GREW, we replace our Baseline with GaitBase-like architecture (4 residual blocks or 10 residual blocks)~\cite{fan2023opengait, he2016deep}, setting channels to (64, 128, 256, 512). More Details are shown in \textit{Supplementary Materials}.
\begin{table}[t]
\centering
\caption{The Rank-1 accuracy (\%) on OccCAISA-B across different views, excluding the identical-view cases.
The NO, CO, SO, and DO denote the testing sets of Non-Occlusion, Crowd Occlusion, Static Occlusion, and Detection Occlusion accordingly.
Based on the walking condition, probe sequences are grouped into Normal Walking (NM), Carrying Bags (BG), and Cloth-changing Condition (CL).
}
\setlength{\tabcolsep}{0.8mm}
\resizebox{\linewidth}{!}{
\begin{tabular}{ccccc|cccc|cccc|cccc}
\toprule
\multirow{2}{*}{Methods} & 
\multicolumn{4}{c}{NO} & \multicolumn{4}{c}{CO} & \multicolumn{4}{c}{SO} & \multicolumn{4}{c}{DO} \\ \cline{2-17}

& NM & BG & CL & Mean & NM & BG & CL & Mean & NM & BG & CL & Mean & NM & BG & CL & Mean \\ \hline

GaitSet~\cite{chao2019gaitset} &92.4 &83.0 &65.2 &80.2 &80.7 &69.6 &50.4 &66.9 &86.0 &76.6 &57.8 &73.5 &85.7 &72.7 &52.7 &70.4 \\

GaitPart~\cite{fan2020gaitpart} &92.3 &84.9 &68.9 &82.0 &80.2 &70.6 &53.3 &68.1 &85.0 &77.3 &60.5 &74.3 &80.7 &68.4 &52.4 &67.2 \\

GaitGL~\cite{lin2021gait} &94.5 &89.2 &75.3 &86.4 &84.4 &74.8 &57.7 &72.3 &87.4 &81.8 &66.9 &78.7 &86.2 &76.2 &61.5 &74.6 \\


STOR~\cite{peng2023occluded} &95.9 &90.9 &77.1 &88.0 &88.8 &80.7 &64.3 &77.9 &91.2 &85.6 &69.9 &82.3 &93.2 &86.3 &70.1 &83.2 \\

GaitBase~\cite{fan2023opengait} &94.4 &88.5 &68.7 &83.9 &87.6 &78.0 &57.5 &74.4 &90.1 &82.7 &62.0 &78.3 &89.9 &80.4 &58.5 &76.3 \\ \hline



\rowcolor{gray!20}\bftab{GaitMoE(\bftab{ours})} &\bftab{96.2} &\bftab{91.5} &\bftab{80.7} &\bftab{89.5} &\bftab{90.0} &\bftab{83.3} &\bftab{68.3} &\bftab{80.5} &\bftab{91.9} &\bftab{86.0} &\bftab{73.9} &\bftab{83.9} &\bftab{93.4} &\bftab{87.3} &\bftab{75.3} &\bftab{85.3} \\
\bottomrule
\end{tabular}
}
\label{OccCASIAB_Performance}
\end{table}
%
\begin{table}[t]
\centering
\footnotesize
 	\caption{
        Comparisons on Gait3D and GREW.
	}
        \setlength{\tabcolsep}{4.0mm}
		\resizebox{0.8\linewidth}{!}{%
			\begin{tabular}{c|c|cc|cc}
				\hline
				\multirow{2}{*}{Method} & \multirow{2}{*}{Venue} & \multicolumn{2}{c}{Gait3D} &\multicolumn{2}{c}{GREW}\\
				\cline{3-6}
				& & Rank-1  &mAP &Rank-1 &Rank-5\\
				\hline
				GaitSet~\cite{chao2019gaitset} & AAAI19 &36.7 &30.0 &46.3 &63.6\\
				GaitPart~\cite{fan2020gaitpart} & CVPR20 &28.2 &47.6 &44.0 &60.7\\
				GaitGL~\cite{lin2021gait} & ICCV21 &29.7 &22.3 &47.3 &63.6\\
				SMPLGait~\cite{zheng2022gait} & CVPR22 &46.3 &37.2 &- &-\\
                MTSGait~\cite{zheng2022gait} & MM22 &48.7 &37.6 &55.3 &71.3\\
				GaitBase~\cite{fan2023opengait} & CVPR23 &64.6 &- &60.1 &-\\
				DANet~\cite{ma2023dynamic} & CVPR23 &48.0 &- &- &-\\
				GaitGCI~\cite{dou2023gaitgci} & CVPR23 &50.3 &39.5 &68.5 &80.8\\
				DyGait~\cite{wang2023dygait} & ICCV23 &66.3 &56.4 &71.4 &83.2\\
				HSTL~\cite{wang2023hierarchical} & ICCV23 &61.3 &55.5 &62.7 &76.6\\ \hline
                \rowcolor{gray!20}\bftab{GaitMoE-T(\bftab{ours})}  &- &\bftab{71.3} &\bftab{62.5} &\bftab{74.4} &\bftab{84.9}\\
                \rowcolor{gray!20}\bftab{GaitMoE-B(\bftab{ours})}  &- &\bftab{73.7} &\bftab{66.2} &\bftab{79.6} &\bftab{89.1}\\
				\hline
			\end{tabular}
			
		}
	\label{gait3d_grew}
\end{table}
%
\textbf{Optimization.} GaitMoE is an end-to-end joint training framework only with ID labels. 
We use the optimizer of SGD with an initial learning rate of 0.1, which is decreasing by a factor of 0.1 per $[10K, 20K, 30K]$, $[10K, 20K, 30K]$, $[20K, 40K, 50K]$, $[80K, 120K, 150K]$ for OccGait, OccCASIA-B, Gait3D and GREW, respectively. 
For joint loss, we set $\beta=0.1$ for OccGait, OccCASIA-B, and $\beta=1.0$ for Gait3D and GREW.
All the models are trained on NVIDIA 8×3090 GPUs.

\subsection{Main Results}
To evaluate the effectiveness of GaitMoE, we first compare with other state-of-the-art methods on OccGait and OccCASIA-B with controlled occlusions, and then further make comparisons on Gait3D and GREW with complex covariates (\eg, uncertain occlusions).

\noindent \textbf{Comparison on OccGait.} 
To qualitatively and quantitatively validate GaitMoE, we build the real-scenario gait database, OccGait, containing a wide range of occlusions and explicit annotations.
Meanwhile, spatial and scale misalignment may occur in all occlusion scenarios.
Tab.~\ref{OccGait_Performance} shows that GaitMoE outperforms other state-of-the-art methods under all types of occlusions.
Notably, CR causes interference from other pedestrians to overshadow the main subject information at certain angles, causing the model to overly focus on the obstructer, not the target. However, the Average results still show SoTA in the main manuscript, which validates the occlusion-solving traits.

\noindent \textbf{Comparison on OccCASIA-B.} Tab.~\ref{OccCASIAB_Performance} shows the overall results where set-based and temporal-based gait recognition methods (\ie, GaitSet, GaitPart, GaitGL, GaitBase) suffer from severe degradation under the occlusion scenario, comparing to their performances in their paper on holistic CASIA-B. 
Although STOR with silhouette registration alleviates the misalignment issue, the complex walking patterns under occlusions make the feature extraction difficult.
Our proposed method outperforms all methods by a large margin, especially in the extremely challenging cloth-changing (CL) condition, \eg, exceeds GaitBase by 12.0\% in \textbf{NO}, and 16.8\% in \textbf{DO}.
The experimental results demonstrate that GaitMoE effectively filters invalid occluded actions and extracts robust actions.

\begin{table}[t]
\centering
\footnotesize
 	\caption{
        Impact of MTE and MAE on OccGait, OccCASIA-B and Gait3D.
	}
        \setlength{\tabcolsep}{1.5mm}
		\resizebox{0.8\linewidth}{!}{
			\begin{tabular}{cc|cccc|cccc|cc}
				\hline
				\multicolumn{2}{c|}{} &\multicolumn{4}{c|}{OccGait} &\multicolumn{4}{c|}{OccCASIA-B} &\multicolumn{2}{c}{Gait3D} \\
				\hline
				 MTE & MAE & NM & CA & CR & ST & NO & CO & SO &DO & Rank-1 & mAP\\
				\hline
				  &  &87.7 &69.0  &75.1  &77.2 &85.8 &75.9  &79.6  &80.0 &59.2 &48.6\\
				 \checkmark &  &89.7 &78.3	&77.3	&81.2	&88.2 &76.9	&82.1	&83.6 &66.2 &56.6\\
				  &\checkmark  &89.4 &78.4 &77.4 &81.8 &87.6 &78.3 &81.7 &82.5 &67.2 & 57.3\\
				 \rowcolor{gray!20}\checkmark & \checkmark  &\bftab{91.4}	&\bftab{82.1} &\bftab{79.9}	&\bftab{84.7}	&\bftab{89.5}	&\bftab{80.5} &\bftab{83.9}	&\bftab{85.3} &\bftab{71.3} &\bftab{62.5}\\
				\hline
			\end{tabular}
			}
	\label{Ab_OccGait_OccCASIAB}
\end{table}
\begin{table}[t]
\centering
\footnotesize
 	\caption{
        The number of Action Experts in MAE on OccGait.
	}
        \setlength{\tabcolsep}{5.0mm}
		\resizebox{0.8\linewidth}{!}{%
			\begin{tabular}{cc|cccc|c}
				\hline
				 MTE & MAE & NM & CA & CR & ST & Mean\\
				\hline
				 \checkmark & 1 & 89.5	&78.3	&78.0 	&82.1	&82.0 \\
				 \checkmark & 8 &90.6	&80.7	&79.1	&83.3	&83.4\\
				\rowcolor{gray!20}\checkmark & 16 &\bftab{91.4}	&\bftab{82.1} &\bftab{79.9}	&\bftab{84.7}	&\bftab{84.5}\\
				 \checkmark & 32 &90.8	&81.2	&79.9	&84.6	&84.1\\

				\hline
			\end{tabular}
			
		}
	\label{Expert_OccGait}
\end{table}

\noindent \textbf{Comparisons on Gait3D and GREW.} We have validated the effectiveness of our method on the controlled occlusion environment (\ie, OccCASIA-B and OccGait). 
The large-scale wild gait databases also provide complex and uncertain occlusion scenarios. 
As shown in Tab.~\ref{gait3d_grew}, GaitMoE-T (4 residual blocks) and GaitMoE-B (10 residual blocks) also achieve the highest performance among state-of-the-art methods, which further proves the generalizability and practicality of our method.

\subsection{Ablation Study}
In this section, we mainly qualitatively and quantitatively validate the MTE and MAE in OccGait, OccCASIA-B and Gait3D. Besides, we provide the visualization of action detection on OccCASIA-B.

\noindent \textbf{The Effectiveness of MTE and MAE.} Tab.~\ref{Ab_OccGait_OccCASIAB} shows that each of MTE and MAE can extract better dynamic information. Especially, the combination of MTE and MAE \ie, GaitMoE, achieves a significant increase over Baseline, which proves that our proposed method discovers the representative actions by first coarse action extraction (\ie, action anchors), and then fine-grained action extraction (\ie, action proposals).
Besides, we select (8, 4) for ($\mathcal S$, $\mathcal K$) on temporal experts since a smaller $\mathcal S$ or larger $\mathcal K$ degrades original information, and a larger $\mathcal S$ or smaller $\mathcal K$ restricts dynamic information. More Details are shown in \textit{Supplementary Materials}.
We also quantify the impact on the number of action experts. Tab.~\ref{Expert_OccGait} illustrates that more experts may result in redundant actions, while fewer experts may not be able to capture the diverse actions.
We select $16$ as the parameter of GaitMoE.

\begin{figure*}[tbp]
	\centering
	\includegraphics[width=1.0\linewidth]{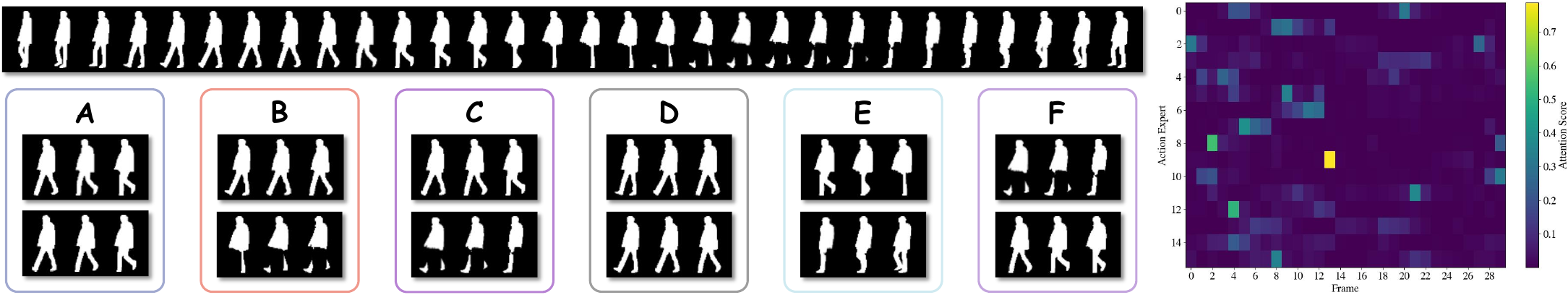}
	\caption{
    The visualization of action composition. Here are an occluded gait sequence, action proposals and the selection map of action anchors. Alphabet denotes the action proposal. Each rectangle denotes one action proposal, and each row within one action proposal denotes an action anchor.
	}
	\label{ablation_action_detection}
\end{figure*}

\noindent \textbf{The Visualization of Action Composition.} 
To better understand and interpret the action detection in gait recognition, we visualize the key module MAE of action detection in Fig.~\ref{ablation_action_detection}, we select part index 60, action anchor with 2 dilated ratios and 6 action proposals as the example. 
%
%
MTE enables to infer the occluded region information by consecutive frames.
For an example in E, the right occluded frame can hallucinate the missing region through the middle frame.
MAE enables to select and integrate the most discriminative similar action anchors from different gait cycles.
For an example in F, although missing information occurs, MAE combines multiple similar action anchors to further confirm this action type (\ie, swinging legs), integrating occluded information.
\begin{figure*}[tbp]
	\centering
	\includegraphics[width=0.8\linewidth]{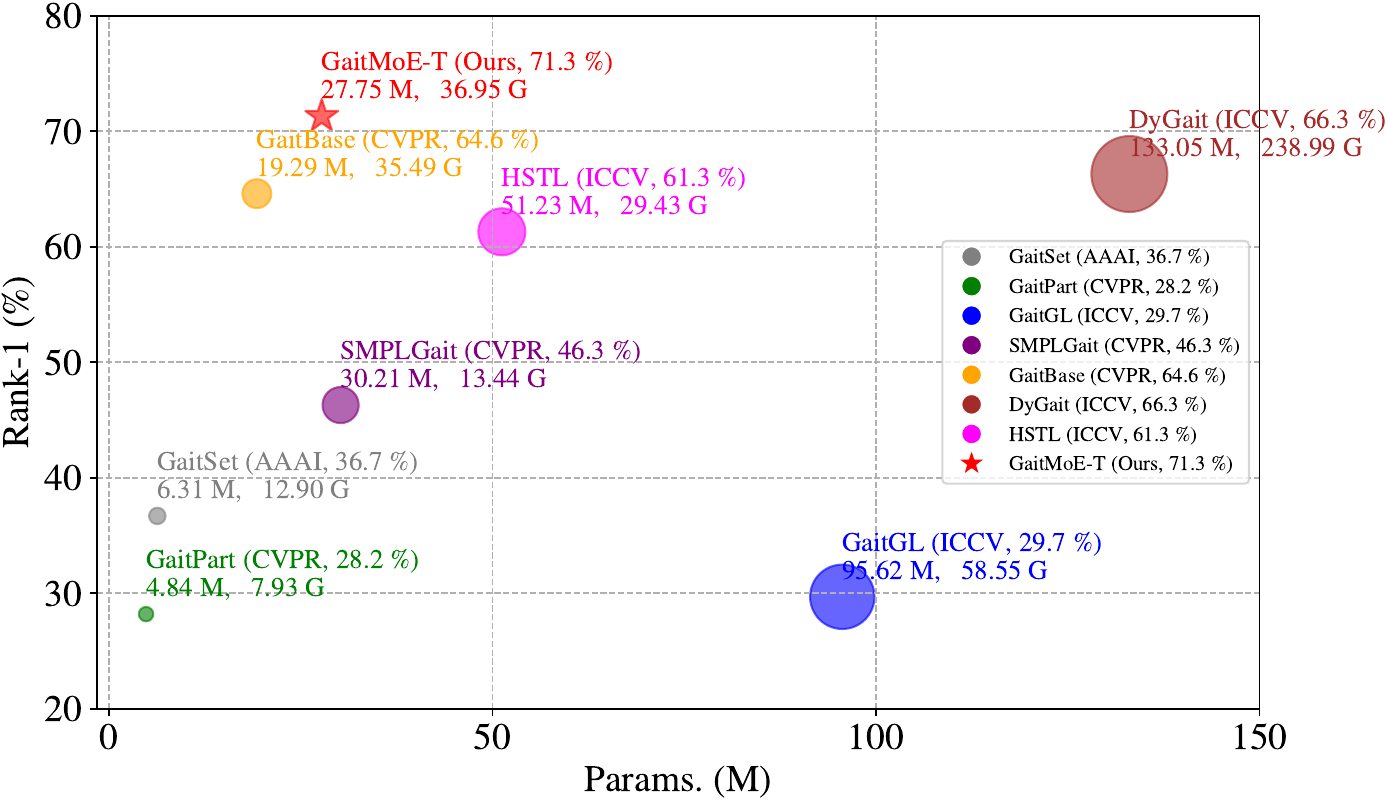}
	\caption{
    The model comparisons on accuracy and efficiency. Rank-1 (\%), Param. (M) and FLOPs. (G) on Gait3D.
	}
	\label{model_comparison}
\end{figure*}

\noindent \textbf{Trade-off between Accuracy and Efficiency.} As Fig.~\ref{model_comparison} shows, we compare all parameters of these models. For FLOPs, models input a 30-frame gait sequence 
without Separate FCs and BNNeck for significant comparisons.
GaitMoE makes a trade-off and achieves SoTA without substantially increasing computational cost. In contrast, DyGait demands significant computation due to 3D convolutions and GaitBase offers better efficiency with 2D convolutions but lower accuracy. 

\subsection{Conclusion and Limitations}
In this paper, we introduce an action detection perspective where a gait sequence is regarded as a composition of actions, allowing holistic body regions to infer occluded body regions and information integration between holistic and occluded actions. 
To detect accurate actions under complex occlusion scenarios, we propose an Action Detection Based Mixture of Experts (GaitMoE) to leverage dynamic contextual information \ie, gait continuity and gait cycle, to construct action anchors and action proposals.
To obtain qualitative and quantitative occlusion analysis, we propose a novel Occluded Gait Recognition benchmark (OccGait) as a pioneering database with a wide range of occlusion scenarios and explicit annotations. 
Extensive experimental results have demonstrated that GaitMoE effectively captures accurate and robust actions for occluded gait recognition.
In addition. we provide some limitations where the selection map of action anchors shown in Fig.~\ref{ablation_action_detection} shows that some of the experts in GaitMoE present redundancy and similarity. 
We will explore the optimization and design for expert selection in the future. 

\section*{Acknowledgment}
This work is jointly supported by National Natural Science Foundation of China (62276025, 62206022), Beijing Municipal Science $\&$ Technology Commission (Z23\break1100007423015) and Shenzhen Technology Plan Program (KQTD2017033109321\break7368).


%
%
\bibliographystyle{splncs04}
\bibliography{main}
\end{document}